\documentclass[letterpaper, 10 pt, conference]{ieeeconf}  % Comment this line out if you need a4paper

\IEEEoverridecommandlockouts                              % This command is only needed if 
\usepackage{commath}
\usepackage{textcomp}
\let\labelindent\relax
\usepackage{enumitem}
\usepackage[ruled,linesnumbered,resetcount,vlined]{algorithm2e}
\usepackage{graphicx} % for pdf, bitmapped graphics files
\usepackage{caption}
\usepackage{subcaption}
\usepackage{multicol}
\usepackage{amsmath}
\usepackage{amssymb}
\usepackage[bookmarks=true]{hyperref}
\usepackage[capitalize,noabbrev]{cleveref}
\usepackage{float}
\usepackage{multirow}
\usepackage{graphicx}
\usepackage{tabularx}
\usepackage[usenames, dvipsnames, table]{xcolor}

\colorlet{mylinkcolor}{black}
\colorlet{mycitecolor}{black}
\colorlet{myurlcolor}{black}

\hypersetup{
  linkcolor  = mylinkcolor,
  citecolor  = mycitecolor,
  urlcolor   = myurlcolor,
  colorlinks = true,
}

\usepackage[markup=bfit, deletedmarkup=sout, authormarkup=superscript]{changes}
\definechangesauthor[name={td}, color=      red] {TODO}
\definechangesauthor[name={bh}, color= NavyBlue] {BH}
\definechangesauthor[name={ar}, color=PineGreen] {AR}
\definechangesauthor[name={ap}, color=Purple] {AP}
\definechangesauthor[name={yt}, color=Orange] {YT}
\definechangesauthor[name={jk}, color=RubineRed] {JK}

\title{\LARGE \bf
\textsl{PokeRRT}: A Kinodynamic Planning Approach for Poking Manipulation
}

\author{Anuj Pasricha$^{*}$, Yi-Shiuan Tung, Bradley Hayes, and Alessandro Roncone%
\thanks{$^{1}$The authors are with the Department of Computer Science,
	     University of Colorado Boulder, 1111 Engineering Drive, Boulder, CO USA
        {\tt\small firstname.lastname@colorado.edu}}%
\thanks{$*$ Corresponding author.}%
}

\begin{document}

\maketitle
\thispagestyle{empty}
\pagestyle{empty}

\begin{abstract}
% \jk{Y'all are all from the same place, so no need for the '1' superscript in the author list. }
This work introduces \textsl{PokeRRT}, a novel motion planning algorithm that demonstrates poking as an effective non-prehensile manipulation skill to enable fast manipulation of objects and increase the size of a robot's reachable workspace. Our qualitative and quantitative results demonstrate the advantages of poking over pushing and grasping in planning object trajectories through uncluttered and cluttered environments.%

% In addition to showing the utility of simulation in the planning loop, Our results both 

% \jk{unachievable?}

\end{abstract}

\section{Introduction}\label{sec:introduction}

% Humans engage naturally in multiple forms of dexterous manipulation that involve grasping, pushing, poking, rolling, and tossing objects \cite{mason2018toward,bullock2011classifying}.
% However, the manipulation skill that has attracted the most attention from roboticists is prehensile manipulation, or \textsl{grasping}.
% Manipulation by grasping is attractive primarily because, once an object is grasped, it generally does not need to be tracked over time and uncertainty on its state is reduced. However, grasping is limited in capability by i) reachability of the robot arm, ii) mechanical design limitations of the end-effector, iii) physical properties of the object being manipulated, and iv) accuracy of the perception system.

\textsl{Non-prehensile manipulation} (NPM) offers a complementary solution to prehensile (\textsl{grasping}) manipulation by significantly expanding the size and dimensionality of the operational space of a robotic manipulator \cite{lynch1999dynamic, zeng2019tossingbot, Huang-1997-14452}. Realistic robot applications such as those that expect the robot to operate in the presence of occlusions, in ungraspable configurations, or in dense clutter, may result in failure modes for robot operation through traditional grasping. Therefore, it is advantageous in such cases to introduce NPM primitives into the robot's skillset.

In this work, we focus on poking as a core NPM primitive and demonstrate how it allows rapid object manipulation and expands the reachable workspace of a manipulator arm. Recent work in NPM has focused on pushing manipulation due to the availability of large-scale datasets \cite{bauza2019omnipush} and the inherent controllability of the skill. Contrary to pushing which operates under the quasistatic assumption to reduce modeling complexity \cite{zito2012two, kloss2020accurate, li2018push}, poking must consider the non-negligible effects of inertial forces since the object continues sliding over its support surface after robot--object contact is broken.

\begin{figure}
\centering
\includegraphics[width=\linewidth]{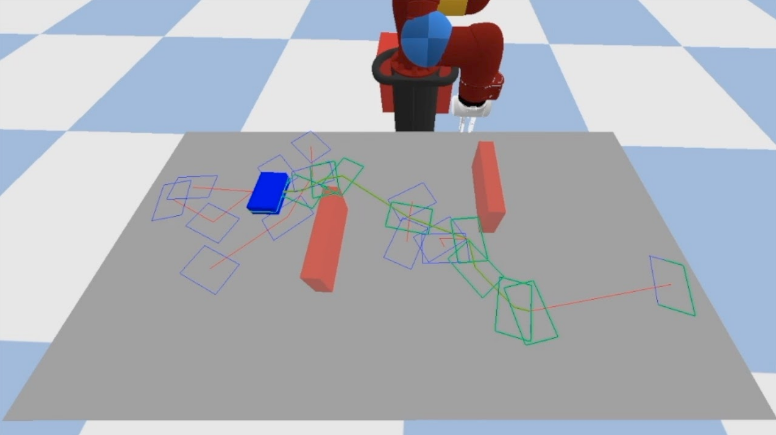}
\caption{\textsl{PokeRRT} plans an object (blue) path using poke actions through an obstacle-rich (red) environment with a kinodynamic approach to sampling-based motion planning. The planning graph is shown in blue and the execution path is highlighted in green.\vspace{-14pt}}\label{fig:scenarios}
\end{figure}
% \section{Background and Related Work}\label{sec:background}

% Research in non-prehensile manipulation dates back to the nineties \cite{lynch1999dynamic} and has focused on skills such as throwing \cite{zeng2019tossingbot}, sliding \cite{shi2017dynamic}, poking \cite{agrawal2016learning,Huang-1997-14452}, and pushing \cite{woodruff2017planning}.

% Past work in pushing manipulation incorporates simulation in a motion planning loop to get to the next feasible state \cite{zito2012two}. Additional contributions in push modeling involve combining object state estimation with affordance prediction from image data to determine contact points for achieving the optimal push \cite{kloss2020accurate} and creating a deep recurrent neural network model to model push outcomes for a variety of objects \cite{li2018push}. However, both approaches use a greedy planner operating in obstacle-free environments. In this work, we propose a sampling-based framework that is capable of planning collision-free paths in the object configuration space. 

% The need for specialized impulse-delivery apparatus to achieve poking is explored in \cite{}. This makes planar manipulation of objects cumbersome due to the manual relocation of the apparatus required. In this work, we use a standard open-chain robot arm to deliver impulses using joint space velocity control.

\section{Current Work}\label{sec:methods}

% \subsection{Formalization of the Poking Motion Primitive}\label{sec:poking}

% \paragraph*{Formalization of the Poking Motion Primitive}

% Poking manipulation is modeled as a process composed of two phases: i) \textsl{impact}, where the robot end-effector makes instantaneous contact with the object; and ii) \textsl{free-sliding}, where the object slides on a planar surface and comes to a stop due to Coulomb friction.
% Two parameters are required to describe the first phase of poking: the point of contact $p_c$ (i.e., where on the contour of the object to strike), and the magnitude of the impact velocity $\|\vec{v}_{EE}\|$.
% Slippage at the contact point between the end-effector and the object may lead to non-linearities; therefore, we fix the direction of $\vec{v}_{EE}$ as being normal to the object's contour.
% Importantly, in order to apply an instantaneous force, the robot must come to a complete halt upon contact with the object. Therefore, collision between the end-effector and the object is treated as an elastic collision. The motor torques applied to stop the end-effector upon contact prevent the impulsive interaction from being truly elastic; however, this can be safely ignored by stopping the end-effector slightly past the contact point.
\textsl{Poking} is an NPM primitive comprised of two phases: i) \textsl{impact}, where the robot end-effector strikes an object at rest and sets it into translational and rotational motion, and ii) \textsl{free-sliding}, where the object slides across a planar support surface and comes to rest due to Coulomb friction.
Poking has a number of desirable characteristics that makes it complementary to grasping \cite{Huang-1997-14452}. Additionally, poking serves as a generalized form of pushing in cases where where applied impulse forces are low.
In this paper, we present a sampling-based kinodynamic planner called \textsl{PokeRRT} which decouples skill modeling and path planning and specifically leverages the following advantages of poking over pushing and grasping: i) it uses instantaneous contact and operates outside the quasistatic regime to expand the size of the manipulator's reachable workspace, ii) it is inherently faster and therefore capable of covering large distances in short periods of time, and iii) it does not impose restrictions on the shape or size of objects being manipulated.
% \subsection{Simulation Model for Poking}\label{sec:sim-model}

% \paragraph*{Simulation Model for Poking}

% In order to understand the effects of impulsive forces on objects, we use the PyBullet physics simulation engine \cite{coumans2019}. The main points of interest when modeling impulsive interactions are contact forces between the robot end-effector and the object of interest, and between the object and the environment (i.e., the object's planar support surface and surrounding obstacles).
% While simulation does not perfectly capture the inherent complexity and stochasticity of real-world contact dynamics due to the simplistic nature of the underlying analytical models used, it nonetheless provide a good balance between pure learning and analytical models by ensuring interaction modeling is both cheap and safe while encapsulating the essential characteristics of robot--object and object--environment interactions.

% \subsection{Motion Planning for Poking}\label{sec:poke-rrt}
% \paragraph*{Motion Planning for Poking}
\begin{table*}[]
\vspace{4pt}
\resizebox{\textwidth}{!}{
\centering
\begin{tabular}{c|c|c|c|c|c|c|c|}
\cline{2-8}
 &
  \multicolumn{6}{c|}{\textbf{Task Time [seconds]}}&\multicolumn{1}{c|}{\textbf{Success Rate}}\\ \hline
\multicolumn{1}{|c|}{\textbf{Planner}} &
  \textbf{S1} &
  \textbf{S2} &
  \textbf{S3} &
  \textbf{S4} &
  \textbf{S5} &
  \textbf{S6} 
  & \textbf{S1 - S6}\\ \hline
% \multicolumn{1}{|c|}{PokeRRT*} & 
% \textbf{46.43 (21.67)} & 
% 212.74 (100.01) & 
% 232.42 (118.63) & 
% 70.47 (38.82) & 
% 132.62 (98.27) & 
% \textbf{130.01 (84.05)} & 
% 0.87 (0.31) \\ \hline
\multicolumn{1}{|c|}{PokeRRT} & 
\textbf{49.69 (21.11)} & 
\textbf{196.54 (124.61)} & 
\textbf{167.55 (138.63)} & 
\textbf{64.14 (52.45)} & 
\textbf{116.35 (89.72)}& 
\textbf{171.77 (103.21)} & 
\textbf{0.88 (0.31)} \\ \hline
% \multicolumn{1}{|c|}{\begin{tabular}[c]{@{}c@{}}Low-Impulse\\ PokeRRT*\end{tabular}} & 
% 169.87 (71.28) & 
% 378.90 (122.21) & 
% 346.70 (111.71) & 
% 165.91 (44.87) & 
% N/A & 
% N/A & 
% 0.53 (0.28) \\ \hline
% \multicolumn{1}{|c|}{\begin{tabular}[c]{@{}c@{}}Low-Impulse\\ PokeRRT\end{tabular}} & 
% 123.71 (36.72) & 
% 328.66 (107.18) & 
% 322.28 (138.05) &
% 135.80 (32.91) & 
% N/A & 
% N/A & 
% 0.63 (0.18) \\ \hline
\multicolumn{1}{|c|}{\begin{tabular}[c]{@{}c@{}}Two-Level\\ Push Planner\end{tabular}} & 
122.68 (47.07) & 
284.78 (104.40) & 
249.32 (68.30) & 
117.58 (67.05) & 
N/A & 
N/A & 
0.44 (0.26) \\ \hline\hline
\multicolumn{1}{|c|}{Pick-and-Place} & 
16.29 (3.76) & 
17.71 (3.81) & 
16.14 (3.25) & 
N/A & 
19.15 (4.99) & 
N/A & 
0.67 (0.00)\\ \hline
\end{tabular}
}
\caption{Task times (shown as \textsl{mean (stddev)}) and success rates (averaged across all six scenarios) are presented for various planning algorithms in simulation across multiple scenarios. Overall, poking is faster than pushing and leads to higher success rates than pushing and grasping.\vspace{-12pt}}
\label{tab:eval-sim-time}
\end{table*}
% This global path planning approach leverages goal and obstacle information in object configuration space to introduce a bias into motion planning and to keep the sampling space low-dimensional to ensure fast planning.
% The proposed planner decouples skill modeling and object path planning to allow for the evaluation of multiple skill models which directly improve planning outcomes.
% The closed-loop nature of our proposed motion planner also compensates for any inaccuracies in simulation modeling while executing poke plans in the real-world. That is, if a resultant pose violates a predefined object pose threshold, we compute a new poke plan from the current object pose to the goal region. This crucial feature allows our planner to plan feasible poke paths in simulation and execute them in the real-world.
% We utilize a PyBullet simulation environment \cite{coumans2019} as the forward physics model in the planning loop to generate impulse-based action paths that respect robot mechanics and represent feasible object motion.

\textsl{PokeRRT} leverages PyBullet as the forward model to validate actions that respect robot and object dynamics \cite{coumans2019}. Path planning for poking operates in the ($x$, $y$, $\theta$) object configuration space in a closed-loop manner to compensate for inaccuracies in the simulation poking model (i.e. it replans if the resultant pose from a poke action is outside a predefined threshold) and consists of six steps:
\begin{enumerate}
    \item Points are sampled uniformly on the object contour and filtered through a conical region originating from a randomly sampled object configuration.
    \item Striking points are generated by extending the sampled contour points away from the object in the normal direction.
    \item Valid end-effector velocity magnitudes that can be applied by the robot are sampled for each striking point.
    \item Feasible actions are applied in simulation to get resultant poses.
    \item The resultant pose closest to the randomly sampled configuration is added to the planning graph.
    \item This procedure is repeated until a resultant pose falls in the task goal region, at which point the shortest path between start and goal object configurations is calculated. The shortest path is defined as one with the lowest overall number of pokes to leverage poking’s capacity to cover large distances quickly.
\end{enumerate}

\section{Evaluation}\label{sec:evaluation}
\vspace{-8pt}
We measure task times (in seconds) and success rates in the final planned path generated by \textsl{PokeRRT} across six test scenarios (see \cref{fig:scenarios}). Task time is defined as the sum of planning, execution, and replanning times. Simulation results are averaged over 250 trials---a trial fails if the planner does not find a valid plan to the goal region in 240 seconds or if the object falls off the its support surface during execution.
To evaluate pushing against \textsl{PokeRRT}, we use the \textsl{Two-Level Push Planner} presented in \cite{zito2012two}. \textsl{Pick-and-Place} is performed in an open-loop manner with predefined grasps for known objects.

\cref{tab:eval-sim-time} shows the task times and success rates for \textsl{PokeRRT}, \textsl{Two-Level Push Planner}, and \textsl{Pick-and-Place}. Poking is successful in all scenarios while pushing fails in S5 and S6 and grasping fails in S4 and S6. \textsl{Two-Level Push Planner} has a low overall success rate ($44\%$) because pushing fails in S5 due to collision with the workspace divider (since pushing, by definition, must maintain constant contact between the end-effector and the object, thereby satisfying the quasistatic assumption). Constant contact also implies that the push action path is longer than that of instantaneous poking, resulting in collisions between the end-effector and workspace obstacles in narrow spaces in S2 and S3. Pushing also fails in S6 due to limited robot reachability. \textsl{Pick-and-Place} always succeeds in simulation for S1, S2, S3, and S5 due to lack of sensing uncertainty. However, it fails in S4 because the object being manipulated is too wide for the gripper and in S6 because the goal pose for the object is out of robot reach.

Task times for \textsl{PokeRRT} are lower than those for \textsl{Two-Level Push Planner} across all scenarios. \textsl{Pick-and-Place} has the lowest task time because it does not involve kinodynamic planning in the object configuration space---the robot moves to object pose, grasps, and moves to goal pose, resulting in a single executed action.

Collectively, our qualitative and quantitative results demonstrate that poking expands robot reachability and dexterity by leveraging instantaneous impact and enabling fast object manipulation through uncluttered and cluttered environments. Our next steps involve extensively testing \textsl{PokeRRT} in the real-world to see if the empirical results support the presented insights from simulation. We also intend to explore additional applications of poking and characterize its dynamics via a combination of learning and analytical models.

\begin{figure}
\centering
\includegraphics[width=\linewidth]{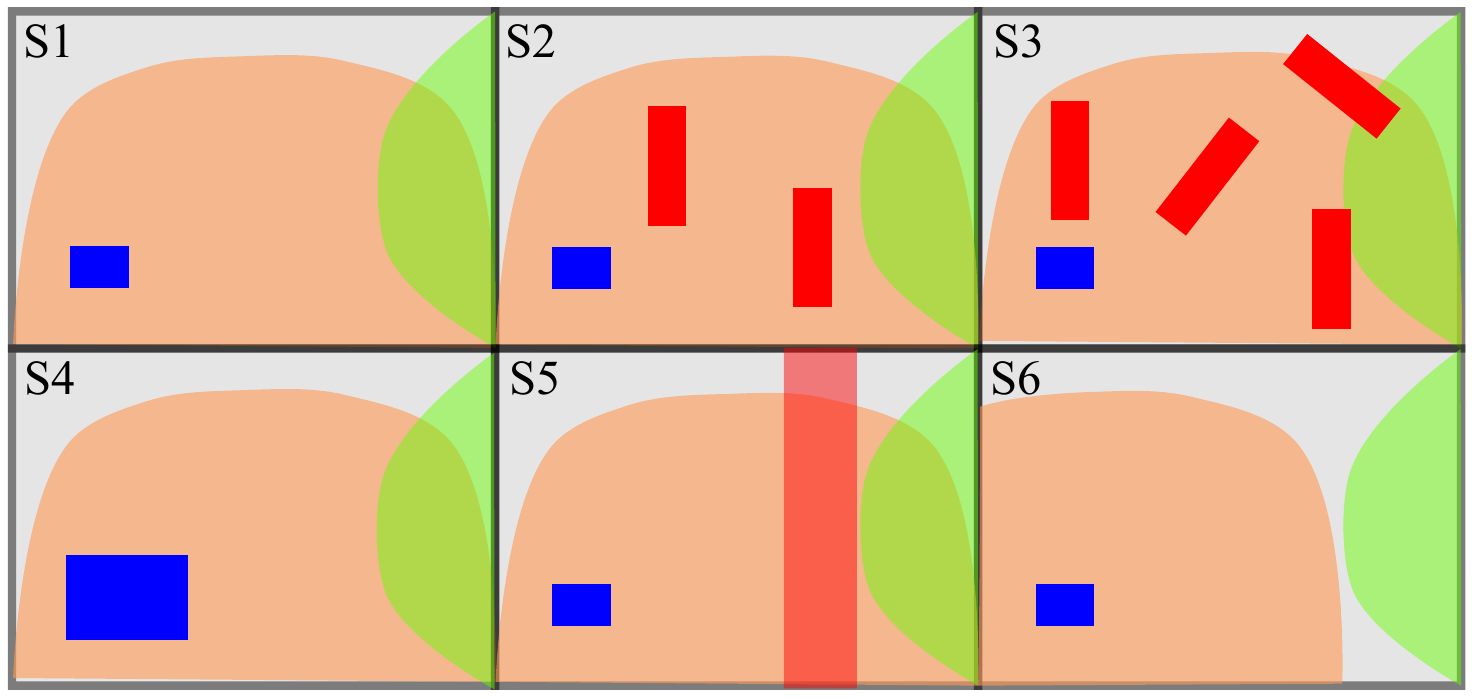}
\caption{\textsl{PokeRRT} is evaluated in six scenarios---no obstacles (S1), $2$ obstacles (S2), $4$ obstacles (S3), wide object (S4), tunnel (S5), and non-overlapping shared workspace (S6). Reachable regions of two robots operating in a shared workspace are depicted in orange and green. The first robot successfully pokes the object (blue) from its reachable workspace (orange) to the goal region (green) in all scenarios while avoiding obstacles (red).\vspace{-12pt}}\label{fig:scenarios}
\end{figure}

% \section{Conclusion and Future Work}\label{sec:discussion}

% Initial results point in favor of poking as a fundamental motion primitive that complements grasping and encompasses pushing in terms of capability.  In addition to characterizing sources of uncertainty in poke modeling, exploring additional applications of instantaneous contact and extending this idea to other non-prehensile manipulation skills are interesting avenues for future work.

% \clearpage
\bibliographystyle{IEEEtran}
\bibliography{iros2021}

\end{document}